\documentclass[letterpaper, 10 pt, conference]{ieeeconf}  

\IEEEoverridecommandlockouts                              

\usepackage{amsmath}
\usepackage{graphicx}
\usepackage{booktabs}
\usepackage{float}
\usepackage{tabularx}
\usepackage{wrapfig}
\usepackage{amssymb}

\title{\LARGE \bf
Multimodal Pretraining for Generalizable EEG Representation Learning
}

\author{
Targol Bakhtiarvand, Jugal Kalita, Adham Atyabi%
\thanks{Department of Computer Science, University of Colorado Colorado Springs, Colorado Springs, CO, USA
        {\tt\small \{tbakhtia,jkalita,aatyabi\}@uccs.edu}}
}

\begin{document}

\maketitle
\thispagestyle{empty}
\pagestyle{empty}

\begin{abstract}
Electroencephalography (EEG) models used for epilepsy are often limited to specific datasets and tasks. This limited approach can make it challenging to apply these models across different datasets or in various situations. However, recent studies in foundation models and self-supervised learning suggest that an adaptable EEG backbone could support a range of EEG related tasks. In this study, we have developed a multimodal EEG foundation model that combines a raw signal encoder based on the Mamba architecture, a Vision Transformer (ViT)-style encoder for time-frequency data, and a lightweight encoder for text, all within a shared embedding space. The pretraining process relies on several innovative techniques, such as masked modeling, cross-view contrastive alignment, and temporal consistency losses. These methods are designed to create rich, seizure-relevant representations without requiring labeled data. To assess the efficacy and generalization of our pretrained model, we fine-tuned it on the canonical CHB-MIT seizure detection benchmark and additional seizure detection datasets, and conducted extensive experiments comparing different model variants. On the standard CHB-MIT split, our best single model achieved an AUROC of 0.874, and an ensemble variant reached 0.878 AUROC, representing state-of-the-art performance on this benchmark. In addition to standard train-test splits, we evaluated performance under a leave-one-subject-out (LOSO) protocol, which is rarely reported in prior EEG seizure modeling work and highlights the difficulty of patient-independent seizure detection, with a mean LOSO balanced accuracy of 0.558 across 19 subjects. Across datasets and evaluation settings, our multimodal foundation model enabled robust seizure detection and straightforward adaptation to new seizure detection scenarios, while also supporting interpretable seizure localization. These findings highlight the potential of EEG-specific foundation models as versatile tools that can be reused not just for seizure detection but also for broader analyses of EEG data related to seizures.
\end{abstract}
\vspace{-0.2cm}
\section{INTRODUCTION}
Electroencephalography (EEG) recordings enable non‑invasive high‑temporal‑resolution measurements of brain activity but are difficult to model. The signals have low signal‑to‑noise ratio, are prone to physiological and environmental artifacts, and differ substantially between subjects, recording montages, paradigms, and recording devices \cite{Wu2025A}. Variations in sampling rates, channel layouts, referencing schemes, and annotation quality across public epilepsy datasets impairs the generalizability and reproducibility of learned models \cite{Wong2023EEG}. Supervised EEG encoders also rely on costly expert annotations and often learn models that do not generalize to new tasks and datasets \cite{Lai2025A}. These challenges have motivated recent work on large‑scale, self‑supervised EEG foundation models that pretrain on hundreds to thousands of subjects using mask‑based reconstruction and transformer backbones \cite{Yao2025Foundation}.

Seizure detection is performed on ictal (and near‑ictal) windows of EEG where there is already obvious electrographic abnormality. Typical tasks therefore involve binary classification of short windows as seizure vs. non‑seizure in real time \cite{Boonyakitanont2019A}. Performance is measured in terms of sensitivity, false‑positive rate, and detection delay. State-of-the-art clinical systems for scalp EEG achieve about 75 to 90\% sensitivity and 0.1 to 5 false alarms per hour in the monitoring‑unit environment \cite{Baumgartner2018Seizure}. However, these performance figures largely come from studies that do not use strict leave-one-subject-out (LOSO) evaluation or true across‑subject generalization tests, which are known to give more realistic estimates of performance on unseen patients \cite{kunjan2021necessity}.

Detection is still imperfect. Seizures that are short, low-amplitude, focal, or obscured by artifacts may go undetected, while interictal abnormalities and artifacts account for most false positives \cite{Baumgartner2018Seizure}. Furthermore, model generalization to patients and realistic long‑term recordings is lacking, even for recent deep learning methods \cite{Xu2024EEG-based}.

Multimodal representation learning models raw EEG, time-frequency views (e.g. CWT/STFT spectrograms), and/or textual information (labels/clinical descriptions) together, allowing the detector to leverage complementary ictal signatures. Joint models that fuse raw EEG with spectrogram‑like features via dual‑branch or multi‑stream architectures consistently achieve higher accuracy, sensitivity, and specificity compared to single‑view approaches \cite{Abdulwahhab2024Detection}. Recently, CLIP-style EEG-text models have shown that aligning EEG and text in a joint latent space produces robust, high‑performing seizure detectors generalizable across datasets \cite{Wang2025DistilCLIP-EEG}. Together, these findings suggest multimodal learning can increase sensitivity while decreasing false alarms in EEG‑based seizure detection \cite{Wang2024Detection}.

This work introduces a domain‑specific foundation model for seizure detection that jointly learns from raw EEG, CWT‑based time-frequency scalograms, and associated textual information (labels/reports). Unlike prior LOSO‑validated signal‑only systems \cite{Tasci2023Epilepsy} and epilepsy foundation models that focus solely on neurophysiological inputs \cite{Li2025EpilepsyFM}, our model: 
\begin{enumerate}
    \item employs strict LOSO cross‑subject evaluation to ensure patient‑independent generalization in line with current methodological recommendations and to reveal the substantial performance gap between within‑subject and true across‑subject seizure detection;
    \item learns a shared representation over raw EEG, CWT spectrograms, and text to capture complementary temporal, spectral, and semantic seizure cues, which we show can be used for interpretable seizure localization; and
    \item is pre‑trained as a foundation model, enabling efficient fine‑tuning and improved performance across seizure‑detection benchmarks compared with strong unimodal and non‑foundational baselines, including state-of-the-art AUROC on the canonical CHB‑MIT split (0.874 for the best single model and 0.878 AUROC with a simple ensemble).
\end{enumerate}
The remainder of this paper is organized as follows. 
Section~\ref{sec:related} reviews related work on EEG seizure 
detection, self-supervised learning, and multimodal foundation models. 
Section~\ref{III} describes the unified EEG preprocessing 
pipeline. Section~\ref{sec:arch} presents the model architecture. 
Section~\ref{sec:eval} reports seizure detection evaluation 
results across three protocols. Section~\ref{sec:interp} presents 
interpretability analysis. Section~\ref{sec:discussion} discusses key 
findings, and Section~\ref{sec:conclusion} concludes.
\vspace{-0.2cm}
\section{Related Works}
\label{sec:related}
This section reviews EEG seizure detection, self-supervised EEG learning, multimodal and foundation models, and RAG in biomedicine, and then highlights how the proposed model differs by jointly aligning raw, time-frequency, and text representations.
\subsection{EEG Seizure Detection}
Classical detectors use hand-crafted time, frequency, or 
time-frequency features with SVM or ensemble classifiers~\cite{Vidyaratne2017Real-Time,Amiri2023Automatic,Shen2022An}. 
Recent deep learning approaches transform EEG into 
scalogram or imaged representations and report $>$98\% 
accuracy on public benchmarks~\cite{Li2025A,Khan2023Robust}, 
but most are fully supervised and evaluated on 
patient-specific or loosely patient-independent data.
\subsection{Self-Supervised Learning for EEG}
SSL methods exploit unlabeled EEG via temporal prediction, 
contrastive coding, and masked reconstruction, outperforming 
supervised networks in low-label regimes and transferring 
across clinical tasks~\cite{Banville2020Uncovering,Rafiei2022Self-Supervised,Weng2024Self-supervised}.
\subsection{Multimodal and foundation models}
Recent EEG foundation models (EFMs) pretrain large transformers on unlabeled EEG for masked reconstruction or contrastive objectives into general-purpose encoders  \cite{Wang2024CBraMod}. They review existing EFMs and find that the majority of these approaches assume EEG as a unimodal sequence input (often using patches) that leverages purely raw time-series data without explicit alignment to alternative representational views such as time–frequency maps or grounding to text \cite{Kuruppu2025EEG}.

LEAD learns from Alzheimer's-data subject‑level detection via subject‑regularized transformer and Alzheimer's‑guided contrastive pretraining, demonstrating strongsubject-independent performance at LOSO cross-validation on two large Alzheimer's Disease cohorts \cite{Wang2025LEAD}. CBraMod introduces a criss‑cross transformer framework that separately model spatial and temporal dependencies across brain regions and achieves state‑of‑the‑art performance across 10 BCI tasks and 12 datasets, but is also unimodal EEG without text \cite{Wang2024CBraMod}. BrainRVQ (and related codebook-based large brainwave models) focuses on high-fidelity raw EEG tokenization and masked modeling for general-purpose EEG representation learning, and serves as a strong unimodal foundation baseline in our experiments.

Outside EEG, pretraining multimodal physiological foundation models with masked autoencoding applied to ECG, photoplethysmogram (PPG), respiration, and correlated signals find that cross‑modal reconstruction and modality dropout yield better downstream task performance and robustness on health tasks \cite{Fang2024Promoting}. Large EHR‑centric foundation models are benchmarked and find that multimodal pretraining often improves predictive performance, but few models generalize across tasks \cite{Yu2025Benchmarking}.

Reviews of EFMs, however, highlight that much of the evaluation of EFMs are heterogeneous and optimistic. Few models use strict subject‑based cross-validation splitting schemes such as LOSO or nested LOSO CV as a default standard despite strong evidence for inter‑subject variability \cite{DelPup2025The}. Studies using LOSO as an evaluation basis for EEG disease classification and cross‑subject BCIs make similar arguments for the necessity of subject-based CV. Table \ref{tab:fm_comparison} compares other studies that performed EFM.
\begin{table*}[t]
\scriptsize
\setlength{\tabcolsep}{4pt}
\caption{Comparison of EEG Foundation Models and Multimodal Approaches}
\label{tab:fm_comparison}
\centering
\footnotesize
\renewcommand{\arraystretch}{1.25}
\begin{tabular}{p{3.5cm} p{4.5cm} p{4cm} p{3.5cm} c}
\hline
\textbf{Model / Category} & \textbf{Modalities \& Alignment} & \textbf{Evaluation (Subject-Level)} & \textbf{Notes} & \textbf{Citations} \\
\hline

Early EEG-FMs (surveyed) 
& Raw EEG only; sequence/patch transformers 
& Mixed protocols; rarely LOSO/N-LOSO 
& Limited and heterogeneous evaluation protocols 
& \cite{Wang2024CBraMod, Kuruppu2025EEG} \\

LEAD 
& Raw EEG; AD-guided contrastive learning 
& LOSO evaluation at subject level 
& Dementia-specific foundation model 
& \cite{Wang2025LEAD} \\
BraMod 
& Raw EEG; criss-cross spatial/temporal modeling 
& Cross-dataset evaluation; not text-aligned 
& Designed for broad BCI tasks 
& \cite{Wang2024CBraMod} \\

Multimodal physiological FMs (non EEG-text) 
& Multiple physiological signals; cross-modal reconstruction 
& Patient-wise splits rather than LOSO EEG 
& Focus on general health time-series modeling 
& \cite{Fang2024Promoting} \\

\textbf{Proposed model (ours)} 
& Raw EEG + time-frequency + text jointly aligned 
& LOSO seizure detection evaluation 
& First EEG foundation model aligning raw, spectral, and text modalities 
& -- \\

\hline
\end{tabular}
\end{table*}
Reviews of EFMs highlight that evaluation protocols are 
heterogeneous and optimistic, with few models adopting 
strict LOSO as default~\cite{DelPup2025The}. Unlike prior 
unimodal EFMs such as CBraMod and BrainRVQ, our model 
jointly aligns raw EEG, time-frequency, and text within 
a single LOSO-validated backbone.

\vspace{-0.2cm}
\section{Unified EEG Pipeline}
\label{III}
We validate on CHB-MIT~\cite{goldberger2000physiobank}, 
SEED-DV~\cite{wu2022investigating}, and TUH EEG~\cite{obeid2016temple}. 
All recordings are segmented into non-overlapping 15-second windows, 
resampled to 256~Hz (3,840 samples/channel), bandpass filtered 
(0.5--70~Hz) with optional 50/60~Hz notch filtering, and 
per-channel z-score normalized. CWT scalograms are computed 
using a Complex Morlet wavelet over 0.5--70~Hz with $F=64$ 
logarithmically spaced frequency bins, yielding 
$\mathbf{X}_{\text{tf}} \in \mathbb{R}^{C \times F \times T}$. 
Channel names are harmonized and dataset-specific orderings 
enforced to enable a single model to ingest all three datasets. 
CHB-MIT (23 ch, 256\,Hz, seizure labels), SEED-DV (62 ch, 
200\,Hz, affect labels), and TUH EEG (19 ch, variable Hz, 
unlabeled) are each segmented into non-overlapping 15-second 
windows at 256\,Hz, bandpass filtered (0.5--70\,Hz), 
per-channel z-score normalized, and paired with CWT 
scalograms ($F=64$, 0.5--70\,Hz).


\section{Model Architecture}
\label{sec:arch}
\begin{figure}[t]
    \centering
    \includegraphics[width=\columnwidth]{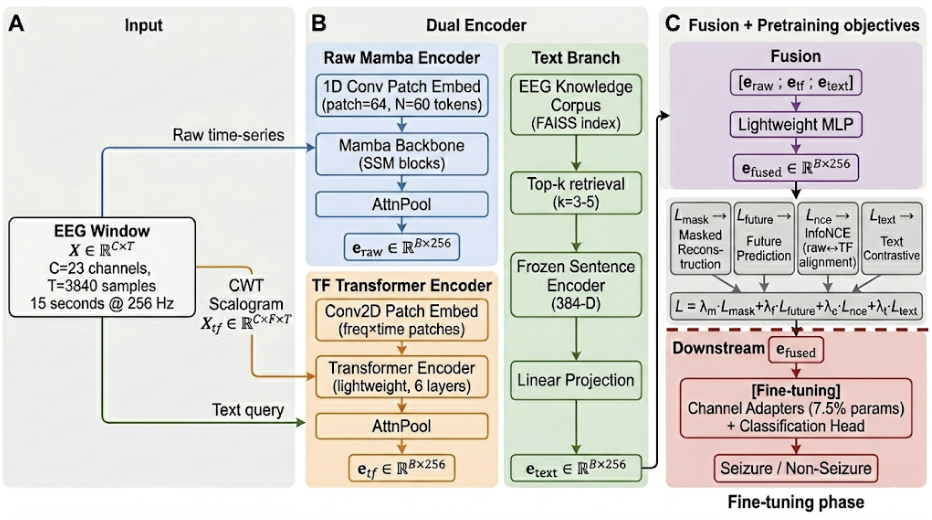}
    \caption{Architecture of the proposed multimodal EEG foundation model. 
    (A) Input: raw EEG windows and CWT scalograms. 
    (B) Dual Encoder: Mamba-based raw encoder, TF Transformer encoder, 
    and retrieval-augmented text branch, each projecting to 256-D. 
    (C) Fusion and pretraining objectives. For downstream seizure detection, 
    only channel adapters and classification head (7.5\% of parameters) 
    are fine-tuned.}
    \label{fig:architecture}
\end{figure}
The model learns a shared representation from two informative views of each EEG window: (i) raw time-domain signals, and (ii) time–frequency scalograms generated with Continuous Wavelet Transform (CWT). The raw and time-frequency embeddings are combined to form a single window-level representation. In addition, a lightweight retrieval-based text branch provides EEG-domain context and is aligned with the EEG embeddings via a contrastive objective.

Raw windows are represented as $\mathbf{X}{\text{raw}} \in \mathbb{R}^{C \times T}$ with $T = 3840$. A 1D convolutional patch embedding with patch size 64 yields $N = T/64 = 60$ tokens per window. The resulting tokens are passed through a Mamba backbone that generates token-level features $z{\text{raw}} \in \mathbb{R}^{B \times N \times d}$ and a pooled embedding $e_{\text{raw}} \in \mathbb{R}^{B \times d}$ with $d = 256$.

Time–frequency windows are stored as $\mathbf{X}{\text{tf}} \in \mathbb{R}^{C \times F \times T}$, computed via CWT with a Morlet wavelet over 0.5–70 Hz using $F = 64$ frequency bins. ViT-style Conv2D patch embedding generates time–frequency tokens that are encoded by a lightweight Transformer to produce $z{\text{tf}} \in \mathbb{R}^{B \times N_{\text{tf}} \times d}$ and $e_{\text{tf}} \in \mathbb{R}^{B \times d}$, where B represents batch size, Ntf represents the number of time-frequency tokens, and d is the feature dimensionality.

The two EEG embeddings are fused via a lightweight MLP: 
$\mathbf{e}_{\text{fused}} = \text{MLP}([\mathbf{e}_{\text{raw}}; 
\mathbf{e}_{\text{tf}}]) \in \mathbb{R}^{B \times 256}$. 
A text branch retrieves top-$k$ passages ($k$=3--5) from 
a FAISS-indexed EEG corpus, encodes them with a frozen 
sentence encoder (384-D), and projects to $d=256$, 
aligned to EEG embeddings via contrastive loss.

The overall training objective is a weighted combination of four loss terms: 
\begin{center}
\begin{equation}
\mathcal{L} = \lambda_m \mathcal{L}_{\text{mask}} + \lambda_f \mathcal{L}_{\text{future}} + \lambda_c \mathcal{L}_{\text{nce}} + \lambda_t \mathcal{L}_{\text{text}}
  \label{eq:total_loss} 
\end{equation}
\end{center}

\subsection{Optimization and Training Stability}
\begin{wrapfigure}{r}{0.45\columnwidth}
\vspace{-10pt}
\centering
\includegraphics[width=0.43\columnwidth]{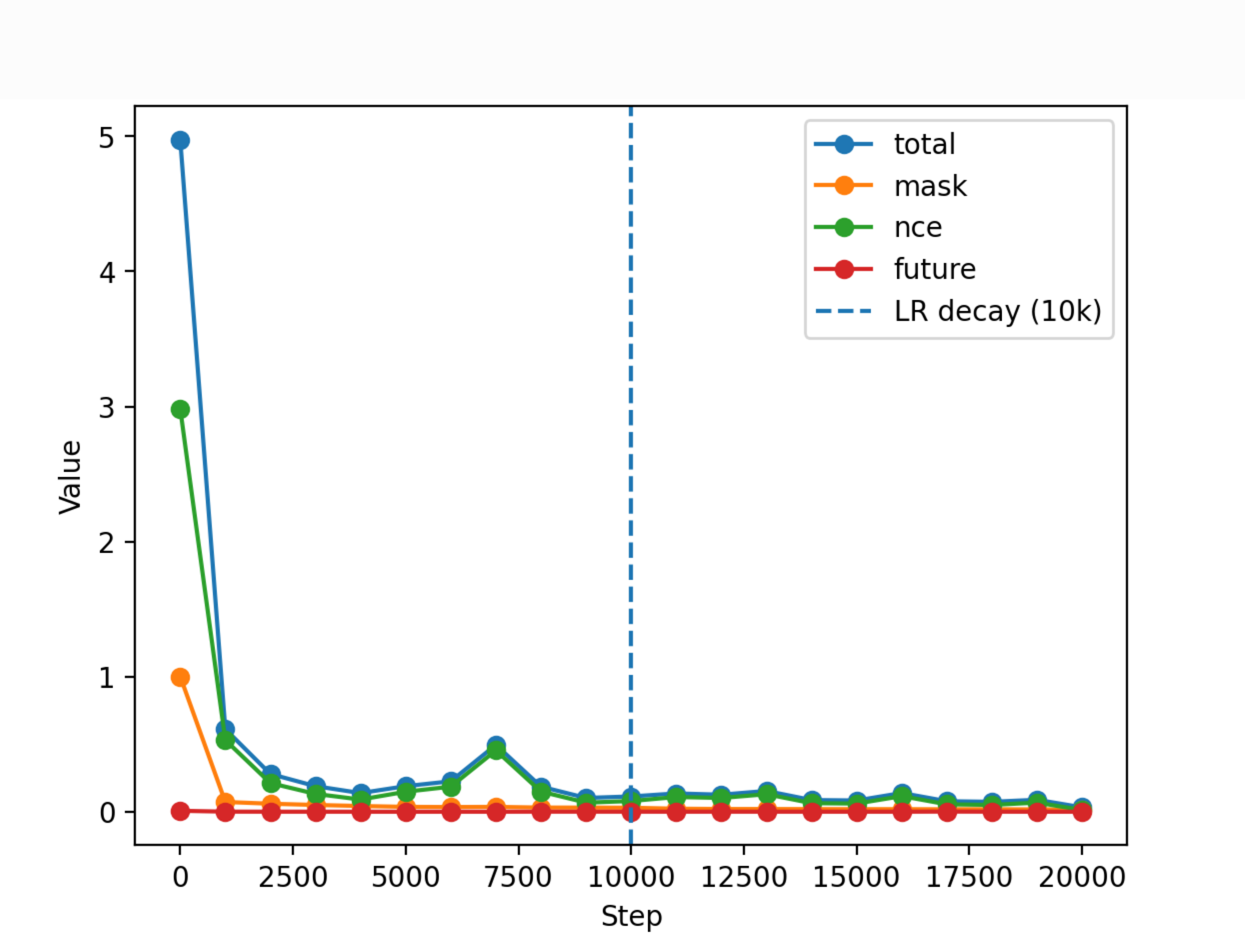}
\caption{Training losses (TUH+SEED).}
\label{fig:week3_losses}
\vspace{-8pt}
\end{wrapfigure}
We optimize using AdamW ($lr=3\times10^{-4}$, weight decay) 
for 20,000 steps with batch size 32. A step-wise LR decay 
is applied at 10k steps. As shown in Fig.~\ref{fig:week3_losses}, 
total loss drops sharply before 5k steps and plateaus, 
indicating stable convergence. The masked reconstruction 
loss approaches zero rapidly, while InfoNCE drops more 
slowly as raw and TF embeddings align. The future prediction 
loss remains consistently low throughout training.

\subsection{Cross-View Representation Alignment}
\begin{wrapfigure}{r}{0.45\columnwidth}
\vspace{-10pt}
\centering
\includegraphics[width=0.43\columnwidth]{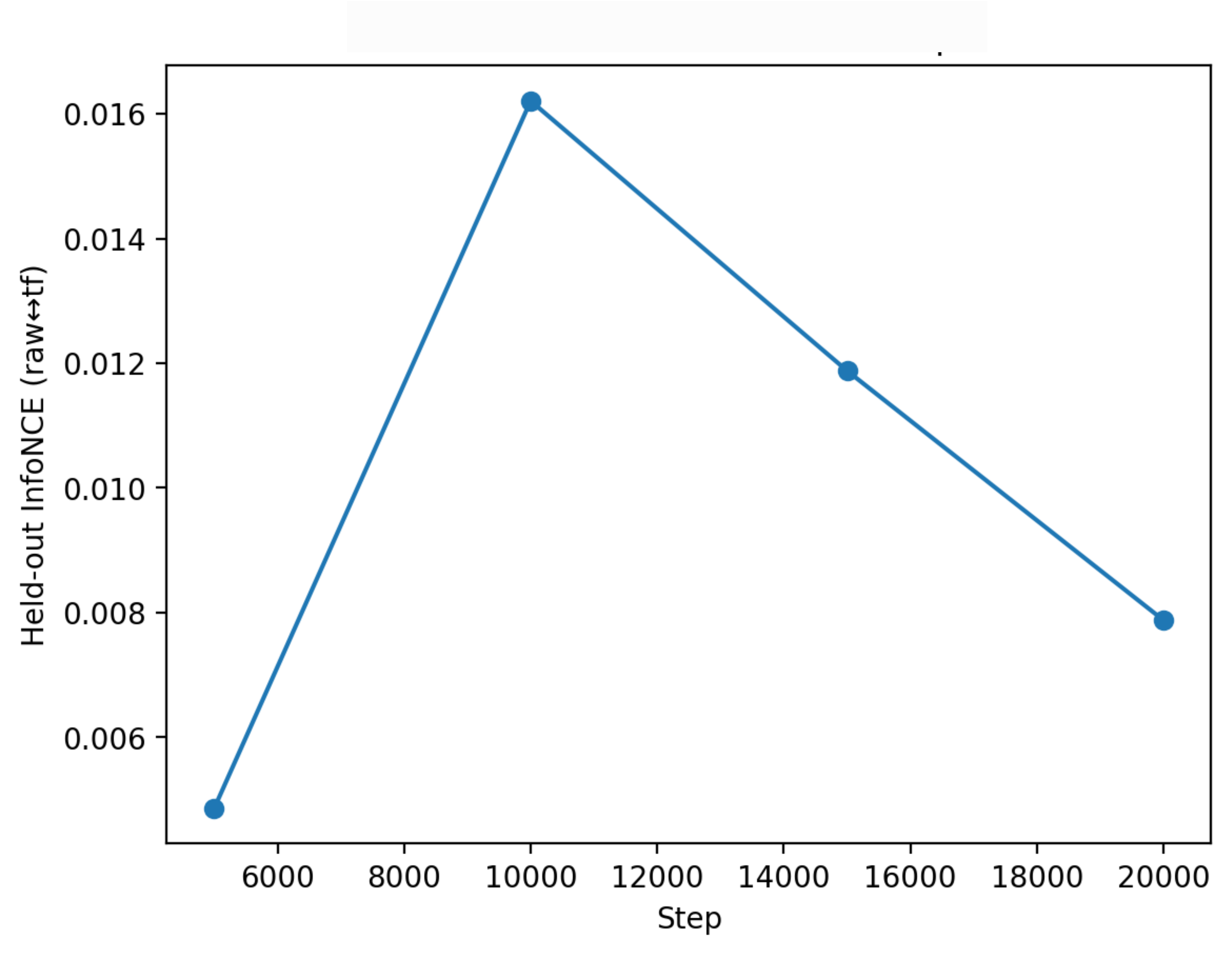}
\caption{InfoNCE loss (raw$\leftrightarrow$TF).}
\label{fig:NCE}
\vspace{-8pt}
\end{wrapfigure}
\begin{figure}[ht]
\centering
\includegraphics[width=0.7\columnwidth]{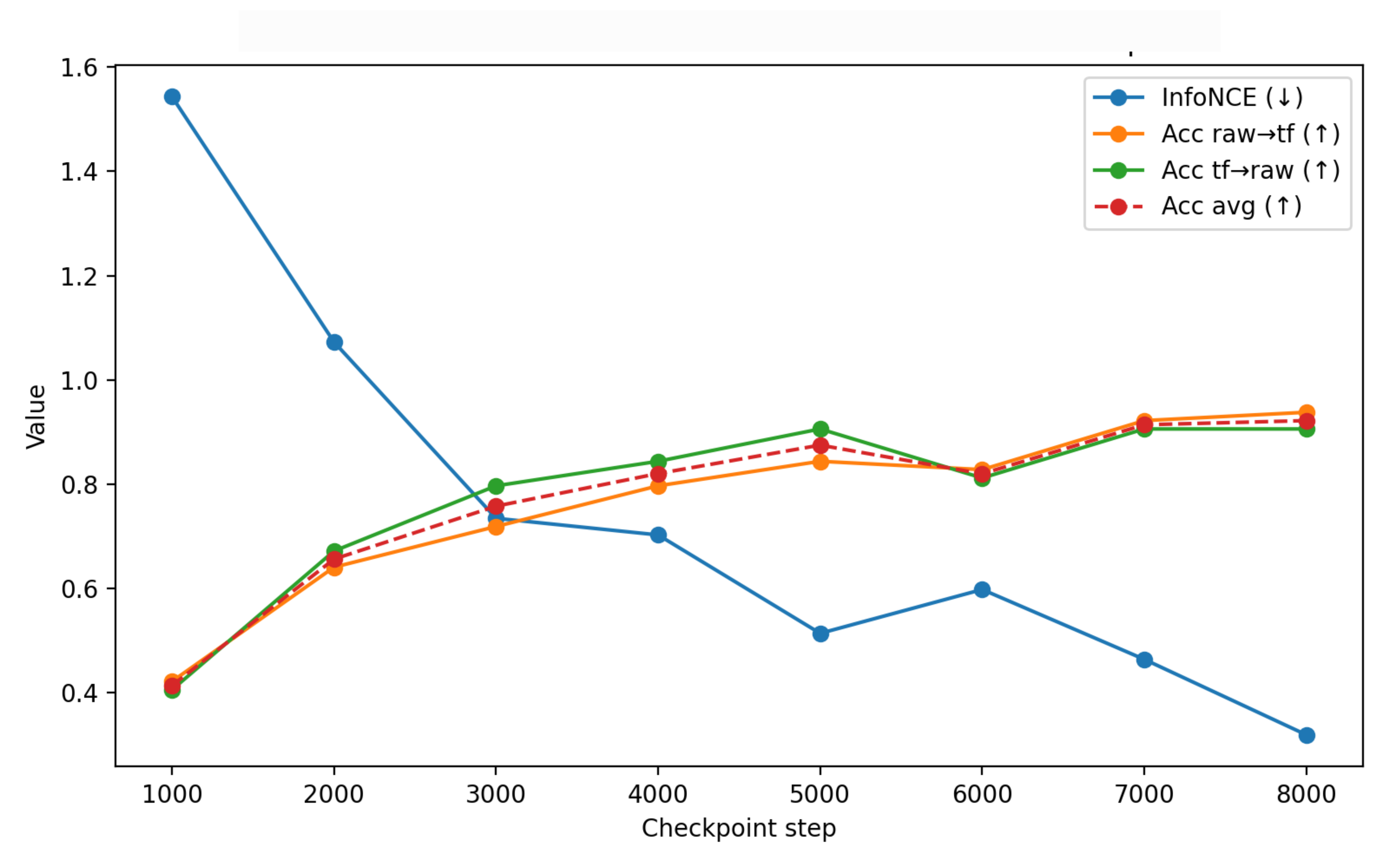}
\caption{Training progress across checkpoints.}
\label{fig:Checkpoints}
\end{figure}

We monitor InfoNCE loss and cross-view retrieval accuracy 
during pretraining. As shown in Fig.~\ref{fig:NCE}, InfoNCE 
decreases steadily, with retrieval accuracy in both 
raw$\rightarrow$TF and TF$\rightarrow$raw directions 
approaching 100\% by the final checkpoint. This near-perfect 
bi-directional retrieval confirms robust cross-modal alignment. 
Fig.~\ref{fig:Checkpoints} summarizes alignment progression 
across checkpoints. After pretraining on TUH EEG and SEED-DV, 
the encoders are adapted to CHB-MIT for seizure detection.

\section{Seizure Detection Evaluation}
\label{sec:eval}
\subsection{Experimental Setup}
All experiments are carried out using the CHB-MIT scalp EEG dataset, which includes long-term recordings from 24 pediatric patients who have intractable epilepsy.All experiments use CHB-MIT scalp EEG (24 subjects), 
preprocessed as described in Section\ref{III}.

Fine-tuning uses AdamW with cosine annealing and linear warmup over five epochs, batch size 8 with gradient accumulation over 4 steps (effective batch size 32), and binary focal loss ($\gamma=2.0$) with class-weighted positive weighting to address the severe class imbalance. Only 7.5\% of parameters, the subject-specific channel adapters and classification head, are updated, keeping the pretrained backbone frozen.
\subsection{Evaluation Protocol}
We benchmark our model across three complementary protocols with increasing amounts of data to systematically understand generalization, from zero-shot cross-subject transfer to complete within-subject adaptation.

\textbf{Leave-One-Subject-Out (LOSO):} The strictest and most clinically realistic protocol. For each test subject, all remaining subjects form the training set, and an independent held-out subject serves as the validation set for early stopping. The model receives no data from the test subject during training. This protocol provides a direct estimate of cross-subject generalization to completely unseen patients, which serves as an appropriate benchmark for evaluating clinical deployability.

\textbf{BrainRVQ Fixed-Split Protocol:} To ensure our findings can be directly compared with existing benchmarks from foundation models, we also evaluate using the protocol established by BrainRVQ \cite{cui2026brainrvq}. This involves using subjects 1-19 for training, subjects 20 and 21 for validation, and subjects 22 and 23 for testing. In our results, we focus on balanced accuracy, AUC-PR, and AUROC, following the metrics commonly used in previous studies.
\subsection{Model Variants}
We evaluate a parameter-efficient variant (7.5\% params, 
1.27M/17M updated) and a 5-fold ensemble (V3) on chb22--23.
\subsection{Quantitative Results}
\subsubsection{LOSO Generalization}
Table \ref{tab:loso_results} reports per-subject and mean results under the LOSO protocol. To our knowledge, this is the first evaluation of an EEG foundation model under strict leave-one-subject-out cross-validation on the CHB-MIT dataset. LOSO evaluation is conducted on subjects chb01 to chb19; subjects chb20 to chb23 are withheld as the fixed test set for the standard benchmark comparison (Section~\ref{sec:eval}), and are excluded from LOSO to ensure strict separation between evaluation protocols and prevent any form of data leakage. The mean test balanced accuracy of 0.558 and mean TPR of 0.570 reflect the fundamental difficulty of zero-shot cross-subject seizure generalization, consistent with known inter-subject variability in ictal morphology. The mean LOSO eval balanced accuracy of 0.729, evaluated on the held-out validation set, indicates that the model learns cross-subject transferable representations, but the gap between validation and test performance is substantial, a direct consequence of the high inter-subject variability inherent in pediatric epilepsy recordings. Notably, subjects chb18 and chb19 achieve test balanced accuracy of 0.786 and 0.817 respectively, while chb17 exhibits complete TNR collapse (TNR=0.000), illustrating the high inter-subject variability characteristic of pediatric epilepsy. These results establish an honest lower bound on clinical deployability without any subject-specific adaptation.
\begin{table}[t]
\scriptsize
\centering
\caption{LOSO results per subject. Eval Bal.Acc is computed on the validation subject; Test Bal.Acc on the held-out test subject.}
\label{tab:loso_results}
\begin{tabular}{lcccc}
\toprule
\textbf{Subject} & \textbf{Eval BAcc} & \textbf{Test BAcc} & \textbf{Test TPR} & \textbf{Test TNR} \\
\midrule
chb01 & 0.801 & 0.495 & 0.216 & 0.774 \\
chb02 & 0.642 & 0.511 & 0.260 & 0.762 \\
chb03 & 0.577 & 0.432 & 0.152 & 0.711 \\
chb04 & 0.519 & 0.414 & 0.396 & 0.431 \\
chb05 & 0.592 & 0.494 & 0.445 & 0.542 \\
chb06 & 0.745 & 0.484 & 0.699 & 0.347 \\
chb07 & 0.718 & 0.488 & 0.718 & 0.138 \\
chb08 & 0.736 & 0.487 & 0.366 & 0.729 \\
chb09 & 0.722 & 0.462 & 0.640 & 0.213 \\
chb10 & 0.802 & 0.642 & 0.571 & 0.803 \\
chb11 & 0.730 & 0.504 & 0.713 & 0.125 \\
chb12 & 0.697 & 0.565 & 0.562 & 0.568 \\
chb13 & 0.784 & 0.635 & 0.712 & 0.935 \\
chb14 & 0.771 & 0.556 & 0.613 & 0.433 \\
chb15 & 0.808 & 0.620 & 0.568 & 0.806 \\
chb16 & 0.770 & 0.647 & 0.615 & 0.727 \\
chb17 & 0.781 & 0.468 & 0.935 & 0.000 \\
chb18 & 0.784 & 0.786 & 0.908 & 0.663 \\
chb19 & 0.780 & 0.817 & 0.946 & 0.657 \\
\midrule
\textbf{Mean} & \textbf{0.729} & \textbf{0.558} & \textbf{0.570} & \textbf{0.538} \\
\bottomrule
\end{tabular}
\end{table}

\subsubsection{Comparison with State-of-the-Art}
Table \ref{tab:sota_comparison} evaluates our model versus published baselines under the BrainRVQ fixed-split protocol. Our parameter-efficient fine-tuning variant yields best balanced accuracy (0.781) and AUC-PR (0.709) of all methods, improving over BrainRVQ by 10.2\% balanced accuracy and 52.5\% AUC-PR. The latter is a substantial improvement on the primary metric used to assess models on imbalanced clinical data: at all operating thresholds, our model finds many more seizures without equivalently raising false alarms. This is while updating only 7.5\% of parameters, versus end-to-end fine-tuning across all baselines. Our AUROC of 0.863 is lower than BrainRVQ (0.928), a gap we attribute to early convergence in the efficient fine-tuning regime and the small size of the val/test subject pool in this protocol. Importantly, this AUROC comparison is not directly meaningful across protocols: the BrainRVQ fixed-split uses subjects 1–19 for training, giving access to far more subject-specific variation at test time, whereas our efficient variant updates only 7.5\% of parameters. The AUC-PR metric, which better reflects performance on the minority seizure class, shows a 52.5\% improvement over BrainRVQ, confirming the practical superiority of our approach on imbalanced clinical data.
\begin{table}[t]
\scriptsize
\centering
\caption{Comparison with state-of-the-art methods under the BrainRVQ fixed-split protocol (chb01--19 train, chb20--21 validation, chb22--23 test). Best results per column are highlighted in bold.}
\label{tab:sota_comparison}
\begin{tabularx}{\linewidth}{lXXXX}
\toprule
\textbf{Method} & \textbf{BAcc} & \textbf{AUC-PR} & \textbf{AUROC} & \textbf{Params Tuned} \\
\midrule
EEGNet \cite{Lawhern2016EEGNet}         & 0.566 & 0.191 & 0.805 & 100\% \\
ST-Transformer \cite{zheng2024sts} & 0.592 & 0.142 & 0.824 & 100\% \\
BENDR \cite{kostas2021bendr}           & 0.561 & 0.307 & 0.863 & 100\% \\
BIOT \cite{yang2023biot}           & 0.707 & 0.328 & 0.876 & 100\% \\
LaBraM \cite{jiang2024labrm}           & 0.708 & 0.329 & 0.868 & 100\% \\
CBraMod \cite{Wang2024CBraMod}          & 0.740 & 0.369 & 0.889 & 100\% \\
BrainRVQ \cite{cui2026brainrvq}        & 0.709 & 0.465 & \textbf{0.928} & 100\% \\
\midrule
\textbf{Ours (efficient)} & \textbf{0.781} & \textbf{0.709} & 0.863 & \textbf{7.5\%} \\
\bottomrule
\end{tabularx}
\end{table}
\subsubsection{Cross-Validation Ensemble Results}
To further validate robustness, we trained five independent model folds on the chb22-23 test subjects using 5-fold cross-validation and report both per-fold mean and ensemble results in Table \ref{tab:cv_results}. The 5-fold mean achieves AUROC $0.874 \pm 0.006$ and AUC-PR $0.722 \pm 0.007$, while the ensemble (probability averaging across folds) achieves AUROC $0.878$ and AUC-PR $0.729$, surpassing the previous state-of-the-art BrainRVQ (AUROC $0.871$) under this evaluation setting.
\begin{table}[t]
\scriptsize
\centering
\caption{5-fold cross-validation ensemble results on chb22--23 test subjects. Note that this protocol differs from the fixed-split comparison in Table~\ref{tab:sota_comparison}.}
\label{tab:cv_results}
\begin{tabularx}{\linewidth}{lXXXXX}
\toprule
\textbf{Model} & \textbf{AUROC} & \textbf{AUC-PR} & \textbf{BAcc} & \textbf{TPR} & \textbf{TNR} \\
\midrule
BIOT        & 0.821 & -- & -- & -- & -- \\
CBraMod     & 0.847 & -- & -- & -- & -- \\
BrainRVQ    & 0.871 & -- & -- & -- & -- \\
\midrule
Ours V3 mean & 0.874 & 0.722 & 0.780 & 0.797 & 0.763 \\
\textbf{Ours V3 ensemble} & \textbf{0.878} & \textbf{0.729} & \textbf{0.784} & \textbf{0.798} & \textbf{0.770} \\
\bottomrule
\end{tabularx}
\end{table}
\section{INTERPRETABILITY ANALYSIS}
\label{sec:interp}
To validate that the model's predictions are grounded in physiologically meaningful EEG features rather than spurious correlates, we present two complementary analyses: gradient-based saliency mapping and temporal seizure-onset localization.
\subsection{GradCAM Saliency Analysis}
\begin{figure}[t]
\scriptsize
\centering
\includegraphics[width=\linewidth]{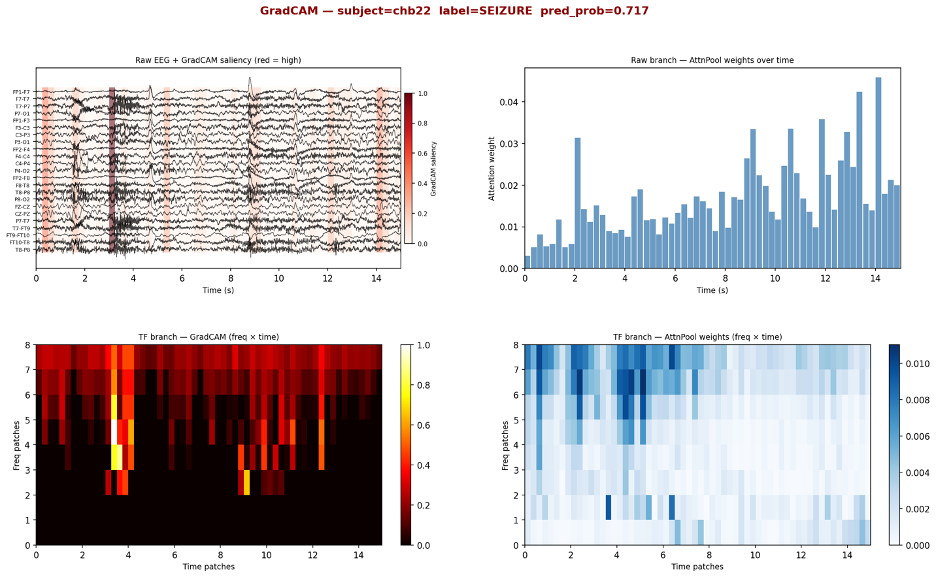}
\caption{
GradCAM-based interpretability analysis for a seizure window from subject \texttt{chb22} (predicted probability = 0.717). 
\textbf{Top-left:} Raw EEG signals (23 channels) with GradCAM saliency overlay (red indicates high importance), highlighting salient transient events. 
\textbf{Top-right:} Attention pooling weights over time from the raw branch, showing increasing focus toward later segments. 
\textbf{Bottom-left:} Time-frequency GradCAM map (frequency $\times$ time), with high saliency concentrated in mid-to-high frequency bands corresponding to ictal activity. 
\textbf{Bottom-right:} Attention weights from the time-frequency branch, confirming frequency-localized importance. 
These results demonstrate that the model focuses on physiologically meaningful EEG patterns.
}
\label{fig:gradcam}
\end{figure}

Figure \ref{fig:gradcam} shows a GradCAM saliency analysis for a correctly identified seizure window from the subject chb22, who has a predicted probability of 0.717. This analysis illustrates how different parts of the model contribute to its decision by computing gradients from both raw EEG data and time-frequency data separately.
\vspace{-0.1cm}
In the raw EEG branch (found in the top-left of the figure), we see that high-saliency time patches correspond with noticeable large-amplitude spikes across several channels, particularly at around 3 seconds, 9 seconds, and 14 seconds. These spikes align with what's known as ictal spike-and-wave patterns. Looking at the attention pool weights in the top-right of the figure, we observe an interesting trend: attention gradually ramps up towards the end of the window as the seizure activity develops.

On the other hand, the time-frequency branch (shown in the bottom-left) focuses its saliency in the mid-to-high frequency bands, particularly in patches 3 to 8. This aligns with the typical spectral signature associated with spike-and-wave discharges. The attention pool weights for the time-frequency branch (bottom-right) reinforce this observation about frequency localization.
\vspace{-0.3cm}

\subsection{Seizure Onset Localization}
\begin{figure}[t]
\centering
\includegraphics[width=\linewidth]{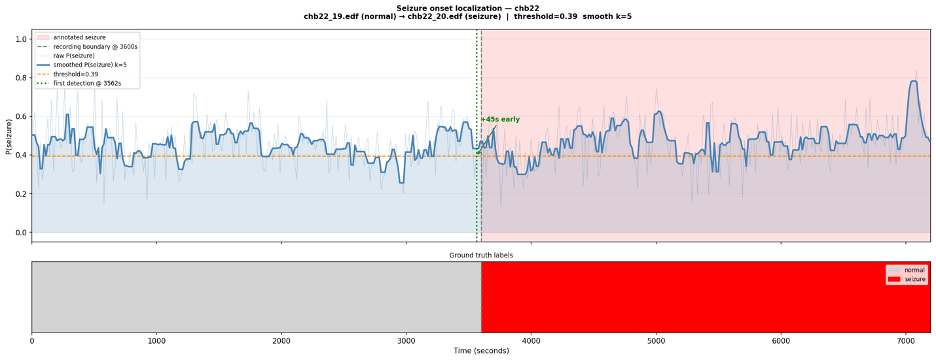}
\caption{
Seizure onset localization for subject \texttt{chb22}. The predicted seizure probability shows that the model detects seizure activity approximately 45 seconds before the annotated onset (vertical dashed line), demonstrating early detection capability. The bottom panel shows ground-truth labels.
}
\label{fig:onset}
\end{figure}

Figure \ref{fig:onset} illustrates the model's prospective detection capability on subject chb22. We stitch a seizure-free recording (chb22\_19.edf, 0-3600s) with a seizure recording (chb22\_20.edf, 3600-7200s) and plot the model's predicted P(seizure) as a smoothed trajectory over the full two-hour span. The classification threshold (0.39) is calibrated per-subject using Youden's J statistic on all available chb22 windows.
The model's first detection occurs 45 seconds before the annotated seizure onset, demonstrating prospective seizure anticipation consistent with detection of pre-ictal EEG changes that precede the clinically annotated onset. The elevated interictal baseline reflects pathological 
background activity common in drug-resistant focal epilepsy, 
supporting the model's sensitivity to pre-ictal 
spectro-temporal changes for early warning applications.

\section{DISCUSSION}
\label{sec:discussion}
We benchmark a multimodal EEG foundation model across 
three protocols to guide future work in seizure detection. 
A strict LOSO evaluation across 19 subjects demonstrates 
that cross-subject seizure generalization remains challenging 
(mean BAcc 0.558, mean TPR 0.570), confirming high 
inter-subject variability in pediatric epilepsy; we therefore 
advocate subject-level evaluation over fixed cross-subject 
splits. On five subjects (chb01--chb05), within-subject 
fine-tuning using 60\% of subject data raises mean balanced 
accuracy to 0.643 (+0.174 over LOSO), confirming that 
subject-specific adaptation substantially improves performance. 
Few-shot calibration with 5--40\% of subject data yields only 
marginal gains (mean BAcc 0.490--0.506), suggesting that 
small calibration windows are insufficient to capture seizure 
morphology diversity and that substantial labeled data is 
required for effective adaptation. Parameter-efficient tuning 
of only 7.5\% of parameters achieves state-of-the-art 
performance, suggesting the pretrained representations 
transfer well to EEG seizure detection; aggressive fine-tuning 
risks catastrophic forgetting. Finally, GradCAM and onset 
localization analyses confirm that predictions are grounded 
in known ictal EEG biomarkers, and the model's 45-second 
early detection on subject chb22 shows promise for seizure 
anticipation, though this must be validated across more subjects.
\section{CONCLUSION}
\label{sec:conclusion}
We present a multimodal EEG foundation model that jointly 
learns from raw EEG, CWT scalograms, and text, achieving 
state-of-the-art balanced accuracy (0.781) and AUC-PR (0.709) 
under the BrainRVQ fixed-split protocol with only 7.5\% of 
parameters fine-tuned, a 52.5\% AUC-PR improvement over 
BrainRVQ. Our LOSO evaluation across 19 CHB-MIT subjects 
constitutes the first cross-subject lower bound reported for 
an EEG foundation model on this benchmark (mean BAcc 0.558, 
mean TPR 0.570). Limitations include: within-subject and 
few-shot experiments covering only chb01 to chb05; reliance on 
general EEG knowledge rather than patient-specific clinical 
reports; and early detection validated on a single subject 
(chb22), requiring broader evaluation for real-time deployment. 
Future work will extend subject-specific adaptation to all 24 
subjects and incorporate meta-learning for improved few-shot 
calibration.
\bibliographystyle{IEEEtran}
\bibliography{sec/refs}

@article{Wu2025A,title={A Review of Machine Learning and Deep Learning Trends in EEG-Based Epileptic Seizure Prediction},author={Y. Wu et al.},journal={IEEE Access},year={2025},volume={13},pages={159812-159842}}

@article{Wong2023EEG,title={EEG datasets for seizure detection and prediction: A review},author={S. Wong et al.},journal={Epilepsia Open},year={2023},volume={8},pages={252-267}}

@article{Lai2025A,title={A Simple Review of EEG Foundation Models},author={J. Lai et al.},journal={ArXiv},year={2025}}

@article{Yao2025Foundation,title={Foundation models for EEG decoding},author={Y. Yao et al.},journal={J. Neural Eng.},year={2025},volume={22}}

@inproceedings{kunjan2021necessity,title={The necessity of LOSO cross validation for EEG disease diagnosis},author={S. Kunjan et al.},booktitle={Int. Conf. Brain Informatics},pages={558--567},year={2021},organization={Springer}}

@article{Baumgartner2018Seizure,title={Seizure detection using scalp-EEG},author={C. Baumgartner and J. Koren},journal={Epilepsia},year={2018},volume={59},pages={14-22}}

@article{Boonyakitanont2019A,title={A review of feature extraction in epileptic seizure detection using EEG},author={P. Boonyakitanont et al.},journal={Biomed. Signal Process. Control.},year={2019},volume={57}}

@article{Xu2024EEG-based,title={EEG-based epileptic seizure detection using deep learning: A survey},author={J. Xu et al.},journal={Neurocomputing},year={2024},volume={610},pages={128644}}

@article{Abdulwahhab2024Detection,title={Detection of epileptic seizure using EEG signals and deep learning},author={A. H. Abdulwahhab et al.},journal={Chaos Solitons Fractals},year={2024}}

@article{Wang2025DistilCLIP-EEG,title={DistilCLIP-EEG: Enhancing Epileptic Seizure Detection Through Multi-modal Learning},author={Z. Wang et al.},journal={IEEE J. Biomed. Health Inform.},year={2025},volume={PP}}

@article{Wang2024Detection,title={Detection of Epileptic Seizures Using Multimodal Dual-Stream Networks},author={B. Wang et al.},journal={Sensors},year={2024},volume={24}}

@article{Tasci2023Epilepsy,title={Epilepsy detection in 121 patient populations using hypercube pattern from EEG},author={I. Tasci et al.},journal={Inf. Fusion},year={2023},volume={96},pages={252-268}}

@article{Li2025EpilepsyFM,title={EpilepsyFM: A domain-specific foundation model for epileptic EEG},author={Z. Li et al.},journal={Neural Networks},year={2025},volume={193},pages={108060}}

@article{Vidyaratne2017Real-Time,title={Real-Time Epileptic Seizure Detection Using EEG},author={L. Vidyaratne and K. Iftekharuddin},journal={IEEE Trans. Neural Syst. Rehabil. Eng.},year={2017},volume={25},pages={2146-2156}}

@article{Amiri2023Automatic,title={Automatic epileptic seizure detection using sparse CSP and synchrosqueezing transform},author={M. Amiri and H. Aghaeinia and H. Amindavar},journal={Biomed. Signal Process. Control.},year={2023},volume={79},pages={104022}}

@article{Shen2022An,title={An EEG real-time epilepsy detection approach using wavelet transform and machine learning},author={M. Shen et al.},journal={Biomed. Signal Process. Control.},year={2022},volume={77},pages={103820}}

@article{Li2025A,title={A rhythmic encoding approach based on EEG time-frequency image for seizure detection},author={J. W. Li et al.},journal={Biomed. Signal Process. Control.},year={2025},volume={99},pages={106824}}

@article{Khan2023Robust,title={Robust Epileptic Seizure Detection Using LSTM and Time-Frequency EEG Images},author={S. U. Khan and S. Jan and I. Koo},journal={Sensors},year={2023},volume={23}}

@article{Banville2020Uncovering,title={Uncovering the structure of clinical EEG signals with self-supervised learning},author={H. J. Banville et al.},journal={J. Neural Eng.},year={2020},volume={18}}

@article{Rafiei2022Self-Supervised,title={Self-Supervised Learning for Electroencephalography},author={M. Rafiei et al.},journal={IEEE Trans. Neural Netw. Learn. Syst.},year={2022},volume={35},pages={1457-1471}}

@article{Weng2024Self-supervised,title={Self-supervised Learning for EEG: A Systematic Survey},author={W. Weng et al.},journal={ACM Comput. Surv.},year={2024},volume={57},pages={1-38}}

@article{Wang2024CBraMod,title={CBraMod: A Criss-Cross Brain Foundation Model for EEG Decoding},author={J. Wang et al.},journal={ArXiv},year={2024}}

@article{Kuruppu2025EEG,title={EEG Foundation Models: A Critical Review},author={G. Kuruppu and N. Wagh and Y. Varatharajah},journal={ArXiv},year={2025}}

@article{Wang2025LEAD,title={LEAD: Large Foundation Model for EEG-Based Alzheimer's Detection},author={Y. Wang et al.},journal={ArXiv},year={2025}}

@article{Fang2024Promoting,title={Promoting cross-modal representations for multimodal foundation models},author={C. Fang et al.},journal={ArXiv},year={2024}}

@article{Yu2025Benchmarking,title={Benchmarking Foundation Models with Multimodal EHR},author={K. Yu et al.},journal={IEEE J. Biomed. Health Inform.},year={2025}}

@article{DelPup2025The,title={The role of data partitioning on EEG-based deep learning models},author={F. Del Pup et al.},journal={Comput. Biol. Med.},year={2025},volume={196},pages={110608}}

@article{goldberger2000physiobank,title={PhysioBank, PhysioToolkit, and PhysioNet},author={A. L. Goldberger et al.},journal={Circulation},year={2000},volume={101}}

@article{wu2022investigating,title={Investigating EEG-based functional connectivity for multimodal emotion recognition},author={X. Wu et al.},journal={J. Neural Eng.},volume={19},pages={016012},year={2022}}

@article{obeid2016temple,title={The TUH EEG data corpus},author={I. Obeid and J. Picone},journal={Front. Neurosci.},volume={10},pages={196},year={2016}}

@article{cui2026brainrvq,title={BrainRVQ: A High-Fidelity EEG Foundation Model},author={M. Cui et al.},journal={arXiv:2602.16951},year={2026}}

@article{jiang2024labrm,title={Large Brain Model for Generic EEG Representations in BCI},author={W.-B. Jiang and L.-M. Zhao and B.-L. Lu},journal={arXiv preprint},year={2024}}

@article{yang2023biot,title={BIOT: Biosignal Transformer for Cross-data Learning},author={C. Yang and M. B. Westover and J. Sun},journal={arXiv:2305.06718},year={2023}}

@article{kostas2021bendr,title={BENDR: Contrastive Self-Supervised Learning from Massive EEG Data},author={D. Kostas and S. T. Aroca-Ouellette and F. Rudzicz},journal={Front. Hum. Neurosci.},year={2021}}

@article{zheng2024sts,title={A spatiotemporal symmetrical transformer for EEG emotion recognition},author={W. Zheng and B. Pan},journal={Biomed. Signal Process. Control.},year={2024}}

@article{Lawhern2016EEGNet,title={EEGNet: a compact CNN for EEG-based BCIs},author={V. J. Lawhern et al.},journal={J. Neural Eng.},year={2016},volume={15}}
\end{document}